\documentclass[lettersize,journal]{IEEEtran}
\usepackage[utf8]{inputenc}

\usepackage{lipsum} 
\usepackage{amsmath,amssymb,amsfonts}
\usepackage{amsthm}
\usepackage{bm}                
\usepackage{mathrsfs}

\usepackage{graphicx}
\usepackage{subfig}            
\usepackage{float}             
\usepackage{booktabs}          
\usepackage{multirow}
\usepackage{makecell}

\usepackage{longtable}
\usepackage{svg}               

\usepackage{algorithm} 
\usepackage{algorithmicx}
\usepackage{algpseudocode}

\usepackage{listings}

\usepackage{cite}
\usepackage{hyperref}
\usepackage{manyfoot}          

\usepackage{textcomp}
\usepackage{stfloats}          
\usepackage{verbatim}

\usepackage{url}

\begin{document}

\title{CLARIFY: A Specialist-Generalist Framework for Accurate and Lightweight Dermatological Visual Question Answering}




\author{Aranya~Saha$^{*}$,~Tanvir~Ahmed~Khan$^{*}$,~Ismam~Nur~Swapnil$^{*}$,~and~Mohammad~Ariful~Haque,~\IEEEmembership{Member,~IEEE}%
\thanks{$^{*}$These authors contributed equally to this work.}%
}

\maketitle

\begin{abstract}
Vision-language models (VLMs) have shown significant potential for medical tasks; however, their general-purpose nature can limit specialized diagnostic accuracy, and their large size poses substantial inference costs for real-world clinical deployment. To address these challenges, we introduce \textbf{CLARIFY}, a \textbf{Specialist–Generalist framework} for dermatological visual question answering (VQA). CLARIFY combines two components: (i) a lightweight, domain-trained image classifier (the Specialist) that provides fast and highly accurate diagnostic predictions, and (ii) a powerful yet compressed conversational VLM (the Generalist) that generates natural language explanations to user query. In our framework, the Specialist's predictions directly guide the Generalist's reasoning, focusing it on the correct diagnostic path. This synergy is further enhanced by a knowledge graph-based retrieval module, which grounds the Generalist's responses in factual dermatological knowledge, ensuring both accuracy and reliability. This hierarchical design not only reduces diagnostic errors but also significantly improves computational efficiency. Experiments on our curated multimodal dermatology dataset demonstrate that CLARIFY achieves an 18\% improvement in diagnostic accuracy over the strongest baseline—a fine-tuned, uncompressed single-line VLM—while reducing the average VRAM requirement and latency by at least 20\% and 5\% respectively. These results indicate that a Specialist–Generalist system provides a practical and powerful paradigm for building lightweight, trustworthy, and clinically viable AI systems.
\end{abstract}


\section{Introduction}
\IEEEPARstart{V}{ision} language models (VLMs) like LLaVA~\cite{liu2023visualinstructiontuning} and Qwen-VL~\cite{bai2023qwenvlversatilevisionlanguagemodel} have demonstrated a remarkable ability to interpret and reason about joint visual and textual data~\cite{zhang2024vision}. Their potential in medicine is vast, with promising applications in tasks ranging from radiological report generation to comprehensive clinical decision support~\cite{chen2023medblipbootstrappinglanguageimagepretraining, tu2024towards}. However, translating this potential into reliable clinical tools faces some critical hurdles. Firstly, their general-purpose pretraining often results in suboptimal performance on specialized diagnostic tasks, a known limitation of applying generalist models to specialized domains like medicine, law, forensics etc~\cite{Wu2023MultimodalLL}. Secondly, their immense size leads to high computational costs and inference latency, rendering them impractical for many resource-constrained clinical environments where on-premise deployment is a necessity~\cite{Santomauro2023EnhancingMI}.

This challenge also propagates to domains like dermatology, a highly visual medical specialty. The accurate diagnosis of skin conditions often depends on subtle morphological features, yet there exists significant visual overlap between different diseases (low inter-class variance) and wide variations in the presentation of a single disease across different skin types and stages (high intra-class variance). This complexity makes dermatological diagnosis a challenging task for general-purpose AI models that lack specialized training and inductive biases for fine-grained visual classification.

Prior attempts to adapt VLMs for such specialized domains have followed two primary paths, each with significant drawbacks. The most common approach is extensive fine-tuning on domain-specific datasets. While this can improve accuracy, it often requires massive labeled datasets and, more critically, can degrade the model's pre-trained conversational and reasoning abilities—a phenomenon known as catastrophic forgetting~\cite{doi:10.1073/pnas.1611835114, luo2025empiricalstudycatastrophicforgetting}. The second path involves post-training model compression to reduce cost, but techniques like pruning~\cite{sun2024simpleeffectivepruningapproach, ma2023llmprunerstructuralpruninglarge} and quantization~\cite{frantar2023gptqaccurateposttrainingquantization, lang2024comprehensivestudyquantizationtechniques} can further impair the model's already fragile diagnostic accuracy. A framework that can simultaneously deliver high specialist accuracy, maintain robust conversational ability, and operate efficiently remains an open and critical challenge.

To resolve this triad of challenges, we propose \textbf{CLARIFY}, a \textbf{Specialist–Generalist framework} designed for dermatological VQA. Instead of relying on a single, monolithic model, CLARIFY adopts a modular, hierarchical approach. It leverages a highly-efficient, domain-specific image classifier—the \emph{Specialist}—to first generate a fast and accurate diagnostic hypothesis. This structured output then guides a compact, conversational VLM—the \emph{Generalist}—to generate a contextually-aware and detailed response. This division of labor ensures that specialized pattern recognition is handled by an expert model, while complex language generation is managed by a VLM that is now primed for the correct diagnostic context.

Furthermore, to enhance the trustworthiness and factual accuracy of the generated explanations, CLARIFY incorporates a retrieval-augmented generation (RAG) pipeline~\cite{lewis2021retrievalaugmentedgenerationknowledgeintensivenlp}. This pipeline utilizes a curated dermatological knowledge graph~\cite{Pan_2024} to retrieve relevant information, grounding the \emph{Generalist's} output in established medical knowledge. This approach makes the system's reasoning more transparent and reliable, directly addressing the critical issue of factual hallucination in medical AI systems~\cite{xiong2024improvingretrievalaugmentedgenerationmedicine}.

We validate our framework on a dermatological VQA task, demonstrating that this synergistic approach leads to a system that is not only more accurate but also significantly more computationally efficient. Our primary contributions are as follows:

\begin{itemize}
    \item \textbf{A Novel Specialist-Generalist Framework:} We propose CLARIFY, an architecture that effectively combines a domain-specific classifier with a generalist VLM to improve both diagnostic accuracy and inference efficiency in medical VQA.
    \item \textbf{A Multimodal Dermatology VQA Dataset:} To facilitate this research, we curated a dataset comprising 1,776  skin disease images from the DermNet website~\cite{dermnet2025}, each paired with multi-turn conversational data also sourced from DermNet, for fine-tuning and evaluation.
    \item \textbf{Knowledge-Grounded Reasoning:} We integrate a knowledge graph-based RAG mechanism that grounds the model's responses in factual domain knowledge, enhancing the system's reliability and interpretability for clinical use.
    \item \textbf{Model Compression for Efficient Deployment:} We have compressed the language model component of the VLM to reduce its size, thereby lowering VRAM requirements and inference costs. Specifically, we applied a structural pruning method to minimize the model's footprint for real-world deployment.
\end{itemize}

\section{Related Works}\label{sec:related_works}

Our research is positioned at the intersection of four key areas: the evolution of Vision-Language Models in medicine, the role of specialized classifiers, the quest for computational efficiency, and the enhancement of model trustworthiness through retrieval-augmented generation.

The landscape of medical AI has been reshaped by the emergence of large, multimodal models capable of processing and reasoning over both visual and textual data. Foundational models like CLIP~\cite{radford2021learningtransferablevisualmodels} and BLIP~\cite{li2022blipbootstrappinglanguageimagepretraining} established powerful methods for aligning image and text representations. This paved the way for conversational VLMs like LLaVA~\cite{liu2023visualinstructiontuning} and generalist medical AI systems such as Med-PaLM M~\cite{tu2024towards}. These models have demonstrated impressive zero-shot capabilities in medical VQA and report generation~\cite{chen2023medblipbootstrappinglanguageimagepretraining}. However, their ``generalist" nature is a double-edged sword: while they possess broad contextual understanding, they often lack the fine-grained discriminative power required for high-stakes diagnostic tasks, leading to critical classification errors~\cite{Wu2023MultimodalLL}. Our work leverages the conversational strengths of these generalist models but argues that they are insufficient for reliable diagnosis on their own.

In contrast to generalist VLMs, traditional ``specialist" models, predominantly based on Convolutional Neural Networks (CNNs) like ResNet~\cite{he2015deepresiduallearningimage} or Vision Transformers (ViTs)~\cite{dosovitskiy2021imageworth16x16words}, have long been the gold standard for medical image classification. These models are trained on domain-specific datasets to excel at a single task, often achieving expert-level performance in areas like dermatology~\cite{esteva2017dermatologist}. Their key limitation, however, is their inflexibility; they produce a class label or a probability score but cannot engage in dialogue, explain their reasoning in natural language, or integrate contextual information from a clinical query. This creates a clear dichotomy: the precision of specialists versus the versatility of generalists.

A major barrier to the clinical adoption of VLMs is their prohibitive computational cost. Deploying models with billions of parameters requires substantial GPU memory and processing power, making on-premise deployment—often a necessity for data privacy—a significant challenge~\cite{Santomauro2023EnhancingMI}. To address this, model compression has become a vital area of research. Techniques like quantization (reducing numerical precision) and pruning (removing redundant model weights) aim to create smaller, faster models with minimal performance loss~\cite{xu2023survey}. Structural pruning, particularly methods that remove entire layers or blocks~\cite{kim2024shortened, men2024shortgptlayerslargelanguage}, offers a practical path to significant inference speedups without requiring specialized hardware. Our framework explicitly incorporates this challenge by using a \emph{compressed} VLM as our Generalist, making efficiency a core design principle rather than an afterthought.

A final challenge for VLMs in medicine is ensuring their outputs are factually accurate and trustworthy. Without explicit grounding in medical knowledge, these models can ``hallucinate" incorrect information. Retrieval-Augmented Generation (RAG) has emerged as a powerful technique to mitigate this by dynamically retrieving information from an external knowledge base to inform the generation process~\cite{lewis2021retrievalaugmentedgenerationknowledgeintensivenlp}. In the medical domain, RAG has been used to enhance clinical question-answering systems by pulling from medical literature or knowledge graphs~\cite{xiong2024improvingretrievalaugmentedgenerationmedicine, zhao2025medragenhancingretrievalaugmentedgeneration}. However, its application within a hybrid VLM framework for diagnostic VQA remains underexplored. Our work integrates a knowledge graph-based RAG to ensure the Generalist's conversational outputs are not only plausible but also clinically sound.

By synthesizing these threads, our work addresses a critical, multifaceted gap: we propose a framework that marries the diagnostic precision of a specialist classifier with the rich, conversational ability of a compressed generalist VLM, all while grounding its output in a factual knowledge base to ensure reliability and trustworthiness.

\section{The CLARIFY Framework}\label{sec:framework}

While large pre-trained models are powerful, specializing them for domains like dermatology or any other highly specialized medical domain is full of challenges. Model compression can address the high computational cost, but adapting these models often leads to a well-documented paradox: fine-tuning for diagnostic accuracy can destroy conversational ability, a phenomenon known as \emph{catastrophic forgetting}. Conversely, maintaining conversational fluency often comes at the cost of clinical precision and factual reliability.

\begin{figure}[h]
\centering
\includegraphics[width=\linewidth]{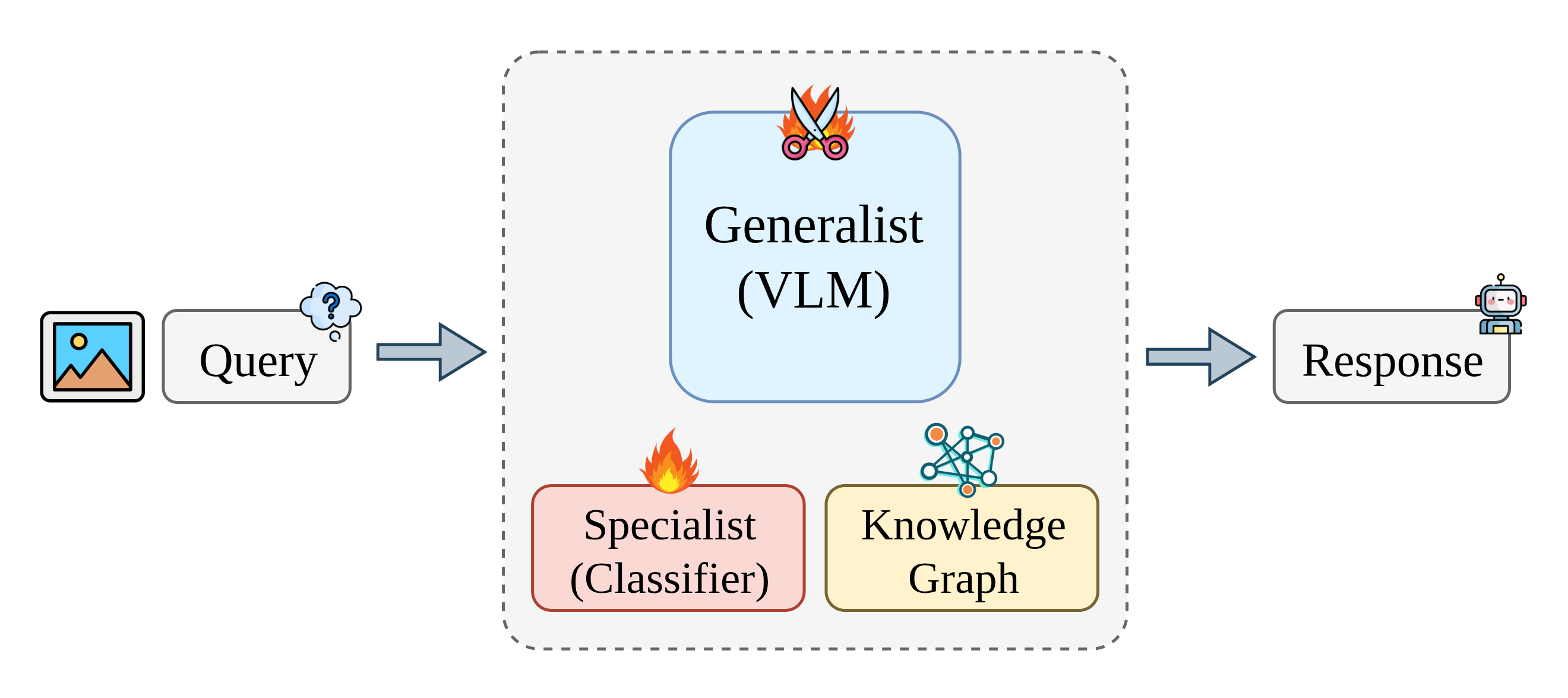}
\caption{High-level architecture of the CLARIFY inference pipeline, demonstrating the decoupling of perception (Classifier) and reasoning (RAG-enhanced VLM).}
\label{fig:clarify_overview}
\end{figure}

To resolve this paradox, we propose the \textbf{CLARIFY} framework, a novel \textbf{Specialist-Generalist} architecture. The high level architecture of the CLARIFY framework is illustrated in Figure~\ref{fig:clarify_overview}.  We named our framework CLARIFY to reflect its core function: bringing clarity to complex medical queries by ensuring responses are both accurate and understandable. Instead of forcing a single model to master two conflicting tasks, CLARIFY decouples the core responsibilities of perception and reasoning into two synergistic components:

\begin{itemize}
    \item \textbf{The Specialist:} A lightweight image classifier finetuned on dermatological images, referred to as the \textit{Specialist (Classifier)} in Figure~\ref{fig:clarify_overview}. Its sole purpose is to perform perception—providing precise and reliable disease identification.
    \item \textbf{The Generalist:} A compressed, general-purpose Vision-Language Model referred to as \textit{Generalist (VLM)} in Figure~\ref{fig:clarify_overview}. It handles reasoning and communication—generating safe, empathetic, and context-aware conversational responses with the help of the integrated knowledge graph.
\end{itemize}

The synergy between these two components is the cornerstone of our framework. The Specialist's predictions are injected into the Generalist's prompt, anchoring its understanding to the correct diagnosis. This guided-prompting mechanism prevents the Generalist from hallucinating an incorrect disease, effectively mitigating the accuracy degradation seen in other approaches. To ensure factual grounding, a knowledge graph-based Retrieval-Augmented Generation (KG-RAG) \cite{sanmartin2024kgragbridginggapknowledge} module provides verified medical context, constraining the Generalist's output to established clinical facts.

In summary, the CLARIFY framework creates a composite system where the Specialist ensures diagnostic precision and the RAG-enhanced Generalist provides safe, factually-grounded explanations. This structure allows the final, compressed model to operate reliably and accurately within a resource-constrained environment, making it suitable for real-world clinical deployment. The following sections will detail the implementation and evaluation of this framework.

\section{Methodology: The CLARIFY System Architecture and Construction}\label{sec:methodology}

\begin{figure*}[h]
    \centering
    \includegraphics[width=\textwidth]{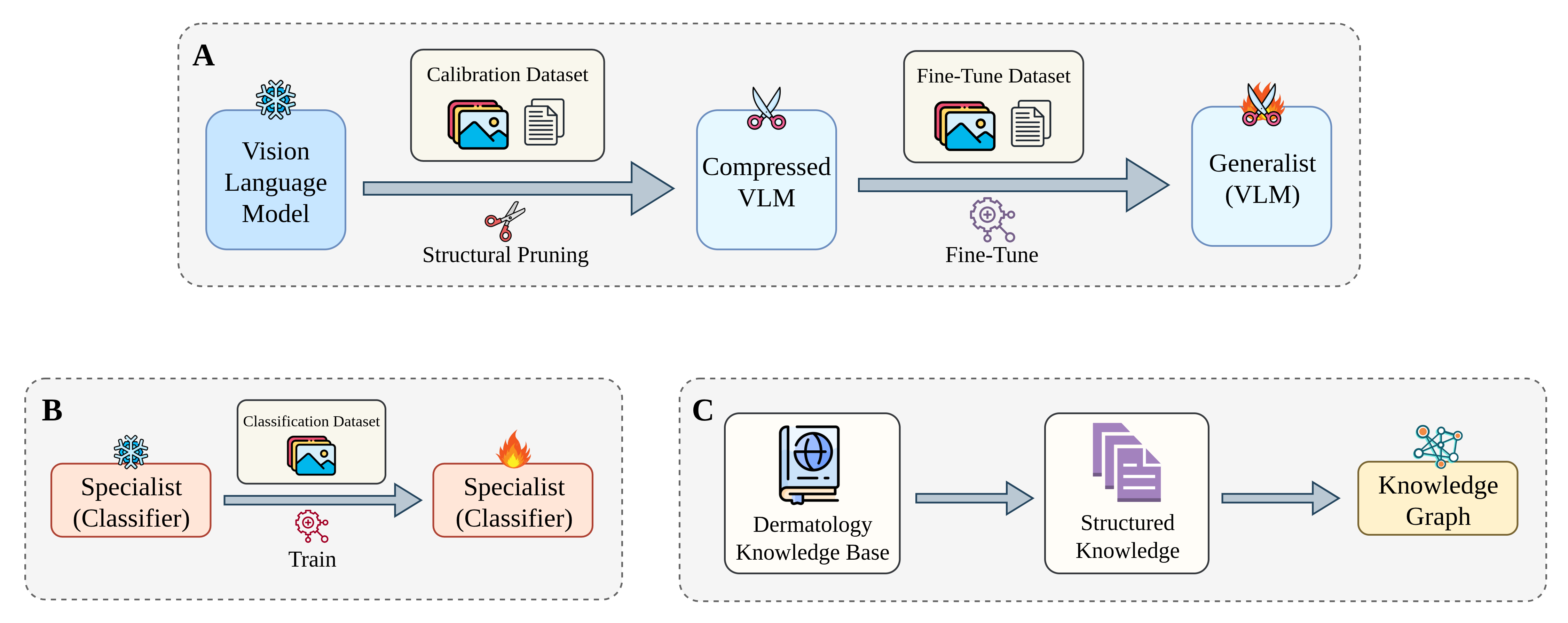}
    \caption{Overview of the structured training and construction flow for the CLARIFY framework: (A) Compression and fine-tuning of a Vision-Language Model (VLM) into a generalist model via structural pruning and fine-tuning. (B) Training specialist classifiers using classification datasets. (C) Integration of dermatology knowledge bases into structured knowledge and conversion into a knowledge graph.}
    \label{fig:training_flow}
\end{figure*}

The CLARIFY framework uses a modular approach that separates different tasks. Instead of building a single, all-in-one model, we create and optimize three separate components: a perception module (the Specialist), a conversational module (the Generalist), and a knowledge module (KG-RAG). This section explains how each component is built and how they work together in the final inference pipeline. Figure~\ref{fig:training_flow} shows the structured flow used to train and build the CLARIFY framework components.

\subsection{Dataset Curation: Small-Derma-VQA}

All experiments were conducted on a curated collection of dermatological images obtained from the DermNetNZ corpus~\cite{dermnet2025}. From the 23 available classes, comprising approximately 19,000 images, we selected eight distinct diseases for our study. We manually vetted the dataset to retain high-quality images belonging to these classes, removed duplicates using Dedup~\cite{gregg2022dedupe}, and constructed a subset of 1,776 images. The selected classes include Actinic Keratosis, Dermatitis, Basal Cell Carcinoma, Rosacea, Melanoma, Psoriasis, Seborrheic Keratosis, and Lichen Planus. Of these, 1,737 images were allocated for training and 39 images were reserved for testing, as summarized in Table~\ref{tab:dataset_distribution}. 

In addition, we extracted question--answer pairs from the same source to construct a complete VQA dataset for our study. We refer to this dataset as \textbf{Small-Derma-VQA}~\cite{aranya_saha_tanvir_ahmed_khan_ismam_nur_swapnil_2025}.

The following are illustrative Question \& Answer (Q\&A) pairs demonstrating typical clinical queries and responses:
\vspace{0.3cm}

\noindent
\colorbox{gray!10}{%
  \parbox{\dimexpr\linewidth-2\fboxsep}{%
    \textbf{Question:} \\ What is the name of the disease? 

    \vspace{0.15cm}

    \textbf{Response:} \\ The name of the disease is Seborrheic Keratosis.
  }%
}

\vspace{0.3cm}

\noindent
\colorbox{gray!10}{%
  \parbox{\dimexpr\linewidth-2\fboxsep}{%
    \textbf{Question:} \\ What are the clinical features of Seborrheic keratoses? 

    \vspace{0.15cm}

    \textbf{Response:} \\ Seborrheic keratoses can appear on most skin areas, except palms, soles, and mucous membranes. They vary in size, color, and texture, from flat to raised, smooth to warty. They may be solitary or grouped on areas like the scalp, breasts, spine, or groin, often sticking to the skin. Ultraviolet (UV) exposure may also contribute to their formation. In some cases, the use of epidermal growth factor receptor (EGFR) inhibitors has been linked to an increased number of these lesions.
  }%
}

\begin{table}[h]
    \centering
    \begin{tabular}{|l|c|c|c|c|}
        \hline
        \multirow{2}{*}{\textbf{Class}} & \multicolumn{2}{c|}{\textbf{Training Set}} & \multicolumn{2}{c|}{\textbf{Testing Set}} \\ \cline{2-5}
        & \textbf{Images} & \textbf{QA Pairs} & \textbf{Images} & \textbf{QA Pairs} \\
        \hline
        Actinic keratosis     & 200 & 2000 & 4  & 40  \\
        Seborrheic keratosis  & 200 & 1000 & 2  & 10  \\
        Melanoma              & 118 & 944  & 4  & 32  \\
        Lichen planus         & 226 & 2260 & 3  & 30  \\
        Rosacea               & 193 & 965  & 2  & 10  \\
        Psoriasis             & 200 & 1000 & 8  & 40  \\
        Basal cell carcinoma  & 300 & 3300 & 9  & 99  \\
        Dermatitis            & 300 & 2700 & 7  & 63  \\
        \hline
        \textbf{Total}        & 1737 & 14169 & 39 & 324 \\
        \hline
    \end{tabular}
    \caption{Class distribution of training and test sets with corresponding number of images and question-answer (QA) pairs.}
    \label{tab:dataset_distribution}
\end{table}

\subsection{The Specialist: A High-Precision Perception Module}

Fine-tuning large-scale vision-language models (VLMs) on custom datasets of limited size is a significant challenge. Generalization is difficult because these models rely on massive and diverse pre-training data. As a result, achieving accurate disease detection with a fine-tuned VLM carries a high risk of overfitting or underperforming. To address this, we designed a dedicated Specialist module, whose sole purpose is to provide the VLM with highly accurate, detection-specific information.

For the core of our Specialist, we selected the DINOv2 model~\cite{oquab2023dinov2}. DINOv2 is a transformer-based model trained via self-distillation on large, unlabeled datasets, which allows it to learn robust and generalizable visual features. Its architecture stacks multiple transformer layers and uses a special `[CLS]' token to aggregate global image information, making it exceptionally effective for downstream classification tasks.

The Specialist is composed of the pre-trained DINOv2 model as a frozen feature extractor and a lightweight classification head. This head is a two-layer feedforward network (FFN) with output neurons corresponding to the number of disease classes in our dataset. 

\textbf{{Mathematical Representation:}} 
Let \(\mathbf{X} \in \mathbb{R}^{H \times W \times C}\) denote an input image of height \(H\), width \(W\), and \(C\) channels. The DINOv2 backbone encodes this image into a latent feature representation:  

\[
\mathbf{Z} = \text{DINOv2}(\mathbf{X}), \quad \mathbf{Z} \in \mathbb{R}^{d},
\]

where \(\mathbf{Z}\) is the \(d\)-dimensional embedding extracted from the final layer of the backbone. This representation is subsequently fed into a trained feed-forward network (FFN):  

\[
\mathbf{y} = \text{FFN}(\mathbf{Z}), \quad \mathbf{y} \in \mathbb{R}^{k},
\]

where \(\mathbf{y}\) contains the unnormalized logits corresponding to the \(k\) disease classes. The final predictive distribution over classes is obtained via the softmax operation:  

\[
\hat{\mathbf{y}} = \text{softmax}(\mathbf{y}),
\]

with \(\hat{\mathbf{y}} \in \mathbb{R}^{k}\) representing the class probabilities.  

\textbf{{Training Strategy:}}
The training process was meticulously staged to maximize performance:
\begin{enumerate}
    \item \textbf{FFN Head Training:} Initially, the entire DINOv2 backbone was kept frozen. We trained only the FFN head using the DINOv2 embeddings as input. This stage used a learning rate of \(1 \times 10^{-3}\) and allowed the classifier to learn the mapping from DINOv2's feature space to our specific disease labels without disturbing the powerful pre-trained weights.
    \item \textbf{Full Model Fine-tuning:} Once the FFN head reached a baseline accuracy of 60\%, we unfroze the DINOv2 backbone and fine-tuned the entire model end-to-end. For this phase, we used a much lower learning rate of \(1 \times 10^{-5}\) with an Adam optimizer and weight regularization. This critical step allows the pre-trained features to adapt subtly to the specific nuances of dermatological images.
\end{enumerate}

\subsection{The Generalist: An Efficient Conversational VLM}
The Generalist module is responsible for reasoning and natural language communication. Its construction involved two key phases: making the base VLM computationally efficient via pruning, and then fine-tuning it exclusively for conversational tasks.

\paragraph{Base Model Compression via Structural Pruning}
Deploying billion-parameter models is computationally expensive. To address this, we applied structural pruning to our base VLM. The language model component of VLMs (e.g. Qwen, LLaVA) contains over 95\% of the total parameters, making it the prime target for pruning. We adopted a coarse-grained structural pruning approach, where entire Transformer layers are treated as individual pruning units. This enables practical acceleration on existing hardware without needing specialized sparse matrix operations.

\begin{figure*}[h]
    \centering
    \includegraphics[width=.49\textwidth]{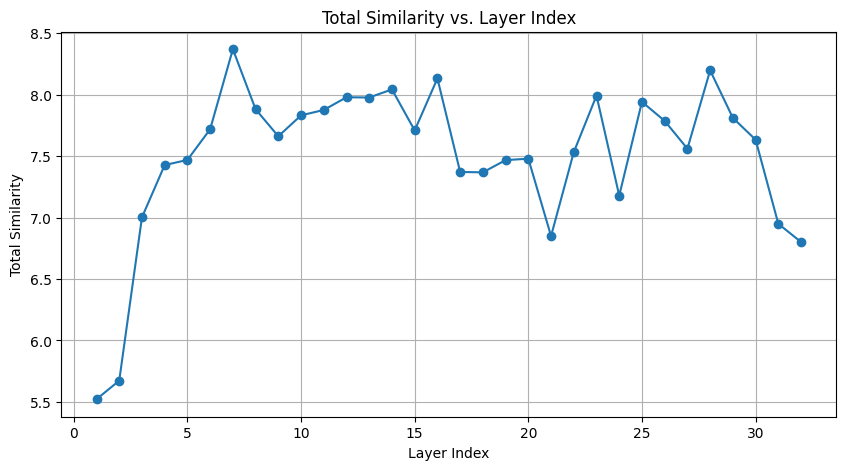}
    \includegraphics[width=.49\textwidth]{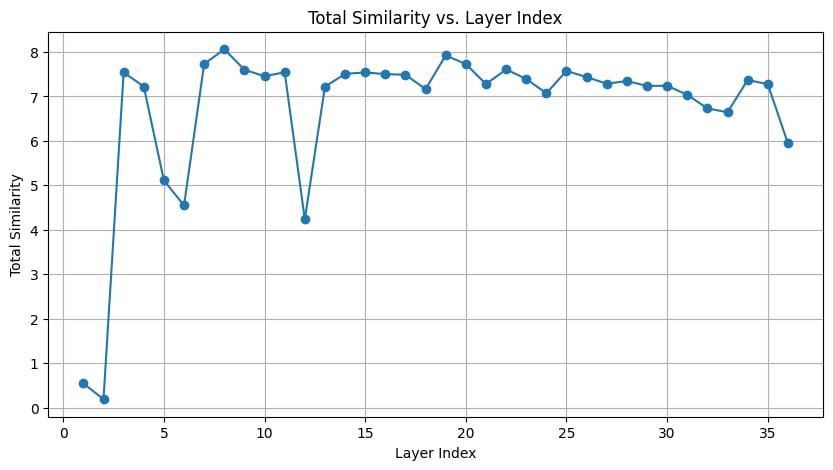}
    \caption{Similarity score across different layers of (left) LLaVA-1.5-7B and (right) Qwen-VL-3B obtained using our pruning analysis. A higher similarity indicates that the model can produce coherent output without the pruned layer, indicating lower layer importance.}
    \label{fig:importance_score_layers}
\end{figure*}

Our pruning strategy, inspired by~\cite{kim2024shortened}, identifies and removes redundant layers. We first established that blocks 1, 2, and 32 are critical and must be preserved~\cite{kim2024shortened, ma2023llm}. For the remaining layers, we employed an iterative, data-driven approach to determine importance. Each candidate layer was temporarily removed, and the pruned model's performance was evaluated on a small calibration dataset, \(\mathcal{D}_{\text{cal}}\).

\begin{figure}[h]
\centering
\includegraphics[width=\linewidth]{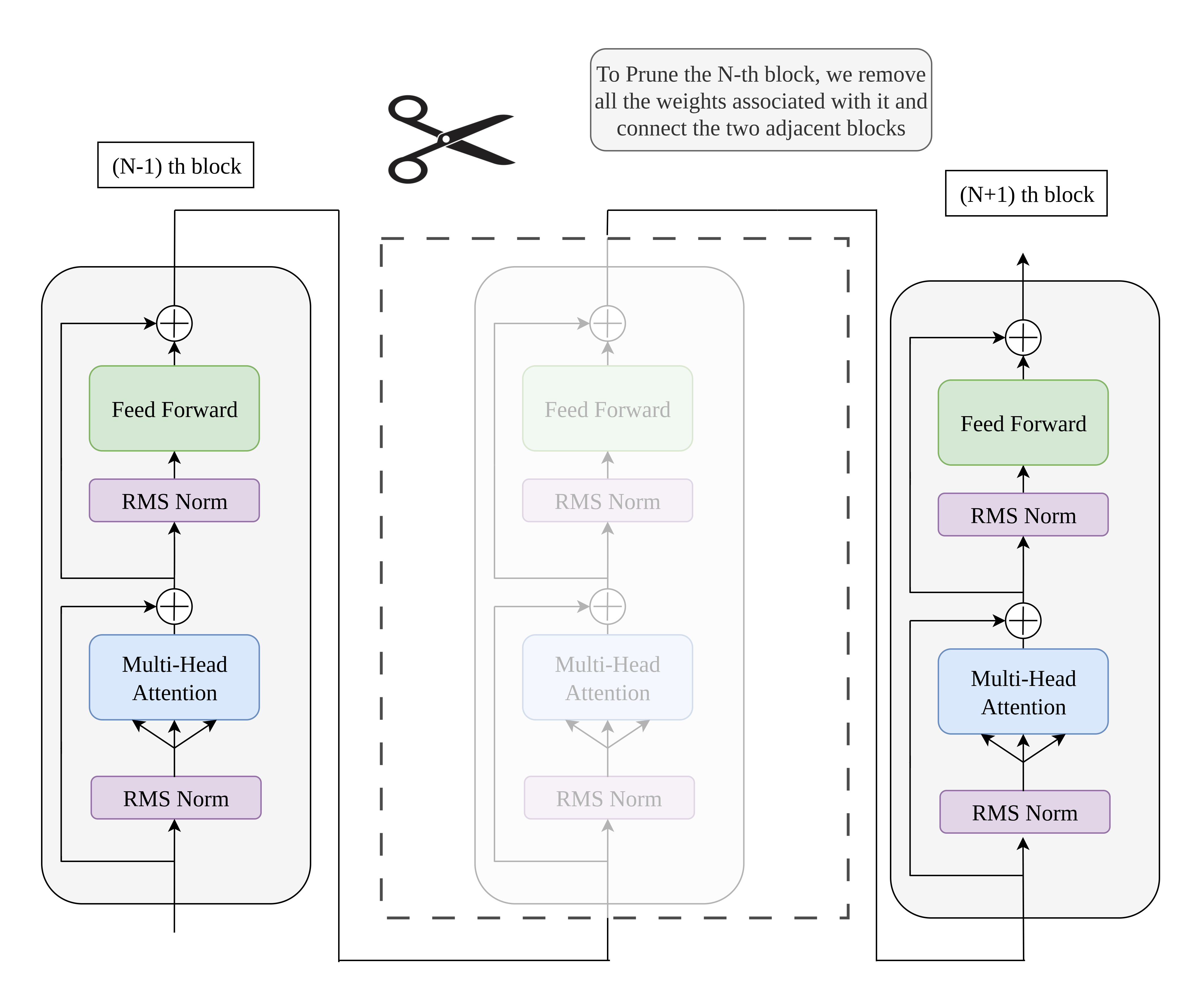}
  \caption{Conceptual Diagram of VLM's Structured Pruning}
  \label{fig:llm_struct_prune}
\end{figure}

To quantify the performance degradation, we compared the output of the pruned model, \(f_{\text{pruned}}\), to that of the original model, \(f_{\text{orig}}\). For each input sample \(x_i \in \mathcal{D}_{\text{cal}}\), we computed the output representations \(y_i^{\text{pruned}}\) and \(y_i^{\text{orig}}\). We then used a pre-trained sentence transformer, \(g(\cdot)\) (all-MiniLM-L6-v2), to get sentence embeddings, \(e_i = g(y_i)\). The cosine similarity between these embeddings served as our metric:
\begin{equation}
    S_i = \frac{ e_i^{\text{pruned}} \cdot e_i^{\text{orig}} }{\| e_i^{\text{pruned}} \| \| e_i^{\text{orig}} \|}
\end{equation}
The average cosine similarity across all samples, \(S_{\text{avg}} = \frac{1}{|\mathcal{D}_{\text{cal}}|} \sum_i S_i\), was taken as the importance score. A higher similarity indicates that the removed layer had a lower impact. Layers with the lowest importance were then permanently pruned. Figure~\ref{fig:importance_score_layers} shows the measured importance scores across different layers.

\paragraph{VLM Specialization for Conversational Reasoning}
To avoid catastrophic forgetting, the compressed VLM was not fine-tuned for disease classification. Instead, its training focused exclusively on its final role: synthesizing information into safe and coherent dialogue. We conducted a single stage of parameter-efficient fine-tuning (using LoRA) on a VQA dataset. The dataset collection strategy is described in Section~\ref{sec:results}. Each sample in the dataset provided the model with a scenario that included a pre-determined disease label (as if from the Specialist) and retrieved facts (as if from the KG), training it to generate an appropriate conversational response. During this stage, the vision encoder and projector were kept frozen, restricting training to the LLM component and aligning its capabilities with its role in the final inference pipeline. This process is illustrated in Figure~\ref{fig:training_flow}, which shows the flow from the Compressed VLM to the Final VLM.

\begin{figure}[h]
    \centering
    \includegraphics[width=\linewidth]{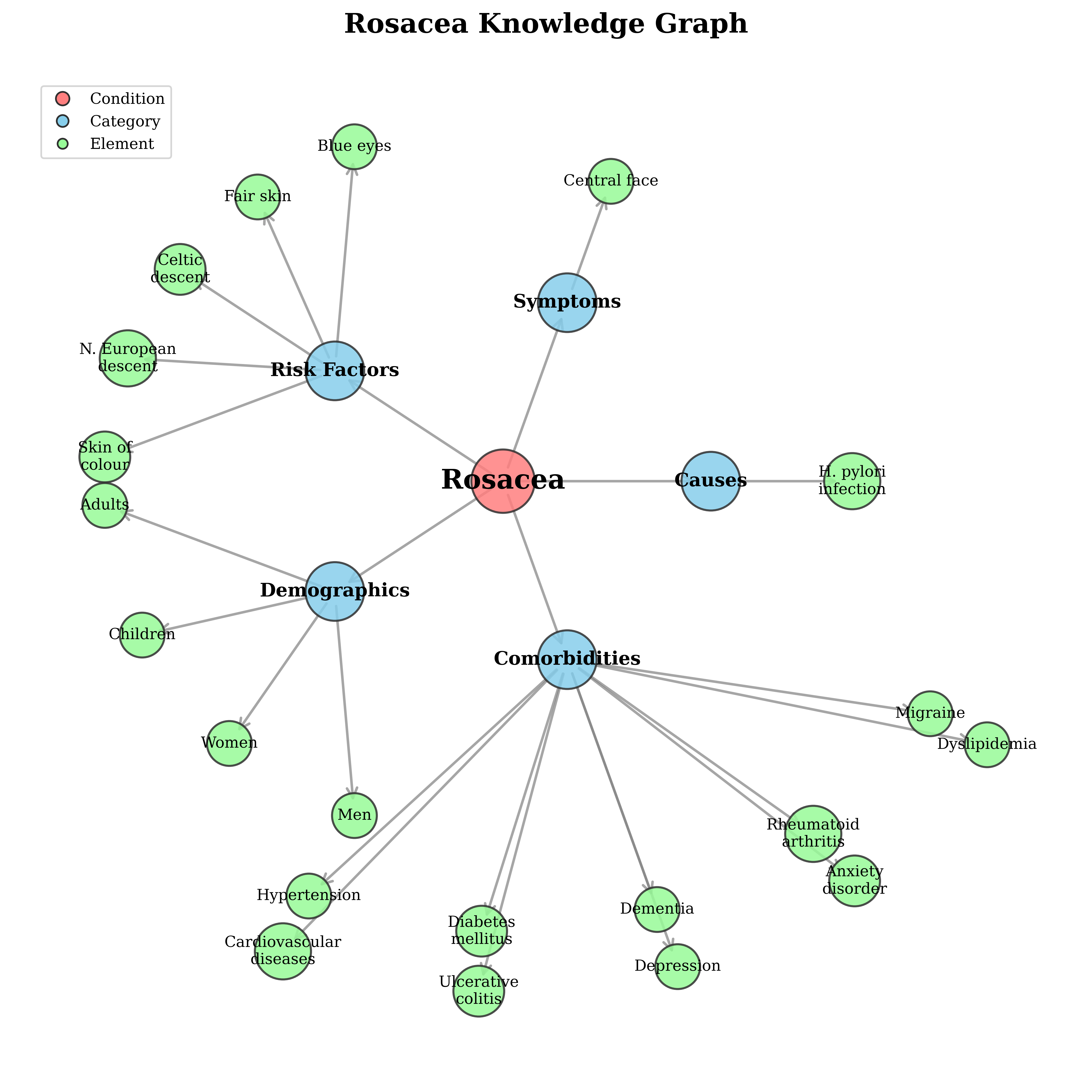}
    \caption{Knowledge Graph example for the disease Rosacea}
    \label{fig:kg_fig}
\end{figure}

\begin{figure*}[h]
  \centering
  \includegraphics[width=\linewidth]{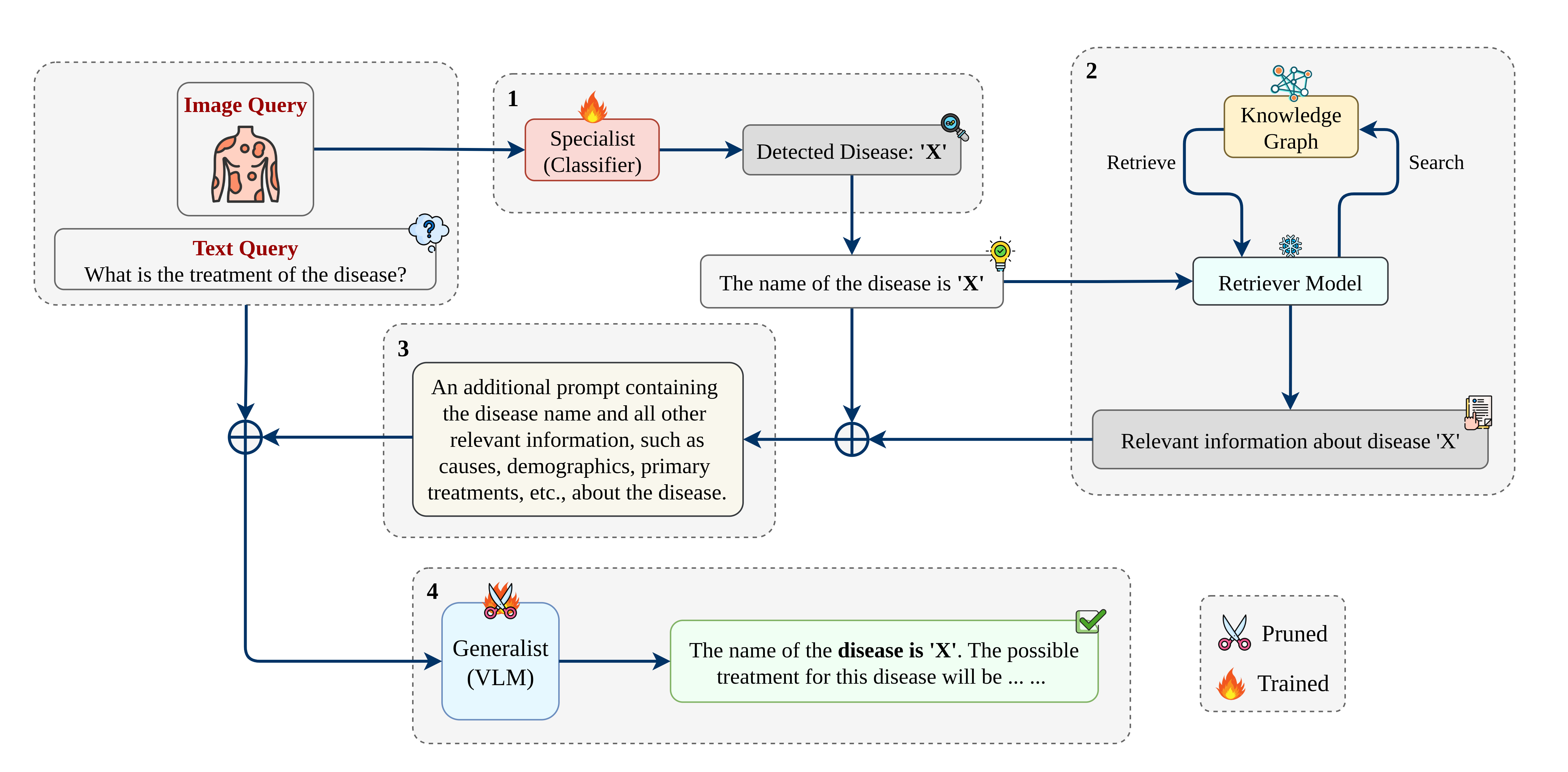}
  \caption{The complete inference workflow of the CLARIFY system. The pipeline shows the sequential flow from the Specialist's prediction, to the KG retrieval, and finally to the Generalist's grounded response generation.}
  \label{fig:dino_llava_kg}
\end{figure*}

\subsection{The Knowledge Module: Grounding Responses with KG-RAG}
To ensure the Generalist's outputs are factually accurate and trustworthy, we integrated a Retrieval-Augmented Generation (RAG) system grounded in a structured knowledge base. RAG enhances generative models by allowing them to retrieve and incorporate external information, thereby improving the quality, accuracy, and relevance of their outputs in practical applications.

\begin{figure}[h]
    \centering
    \includegraphics[width=\linewidth]{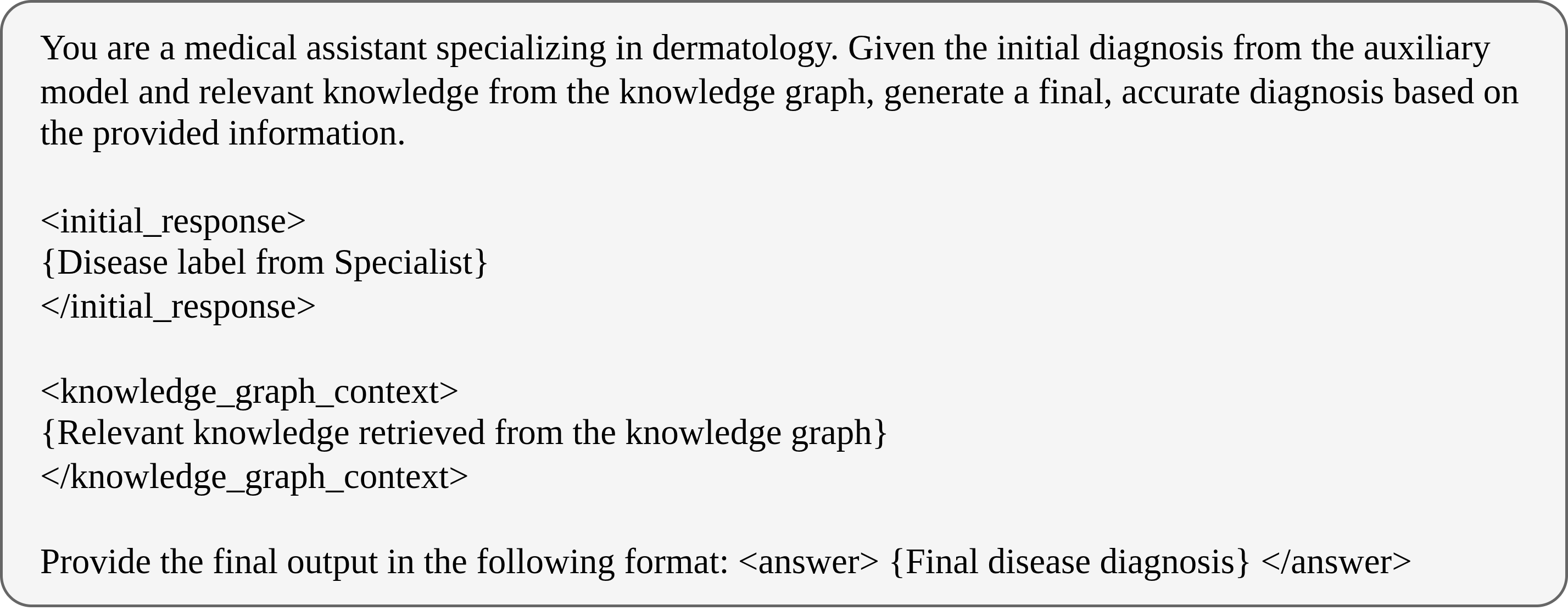}
    \caption{Final prompt for the VLM, constructed using the detected disease information from the specialist module along with additional context from the knowledge graph.}
    \label{fig:prompt}
\end{figure}

We constructed a dermatological knowledge graph (KG) using the KGGen platform~\cite{mo2025kggen}, which employs large language models to extract entities (diseases, symptoms, treatments) and their relationships. For our implementation, we used Wikipedia as the source for entity extraction, acknowledging that it is not a clinically authoritative database, but it serves as a scalable and diverse resource for prototyping. The KG was designed to demonstrate the feasibility of knowledge-grounded reasoning in our system and to illustrate how structured information can reduce hallucinations in language model outputs. The knowledge database is first passed to an LLM, which in our case is Gemini 2.0~\cite{geminiteam2025geminifamilyhighlycapable}, to generate an initial knowledge graph. The LLM is then called a second time through DSPy~\cite{khattab2023dspy} to verify the graph in context, thereby improving its consistency and reducing redundancy in entity-relation pairs. To enable efficient semantic matching, all entities and relations in the KG were encoded into vector embeddings using the all-MiniLM-L6-v2 sentence transformer~\cite{wang2020minilmdeepselfattentiondistillation}, which offers a balance of speed and performance. The framework is agnostic to the underlying knowledge source; replacing Wikipedia with clinically validated ontologies would directly enhance the reliability, precision, and clinical relevance of the system. A sample conceptual diagram of a knowledge graph for the disease rosacea is shown in Figure~\ref{fig:kg_fig}, which highlights how key entities such as symptoms, risk factors, and treatment options are interconnected and can guide more grounded responses.

\begin{figure*}[h]
  \centering
  \includegraphics[width=\linewidth]{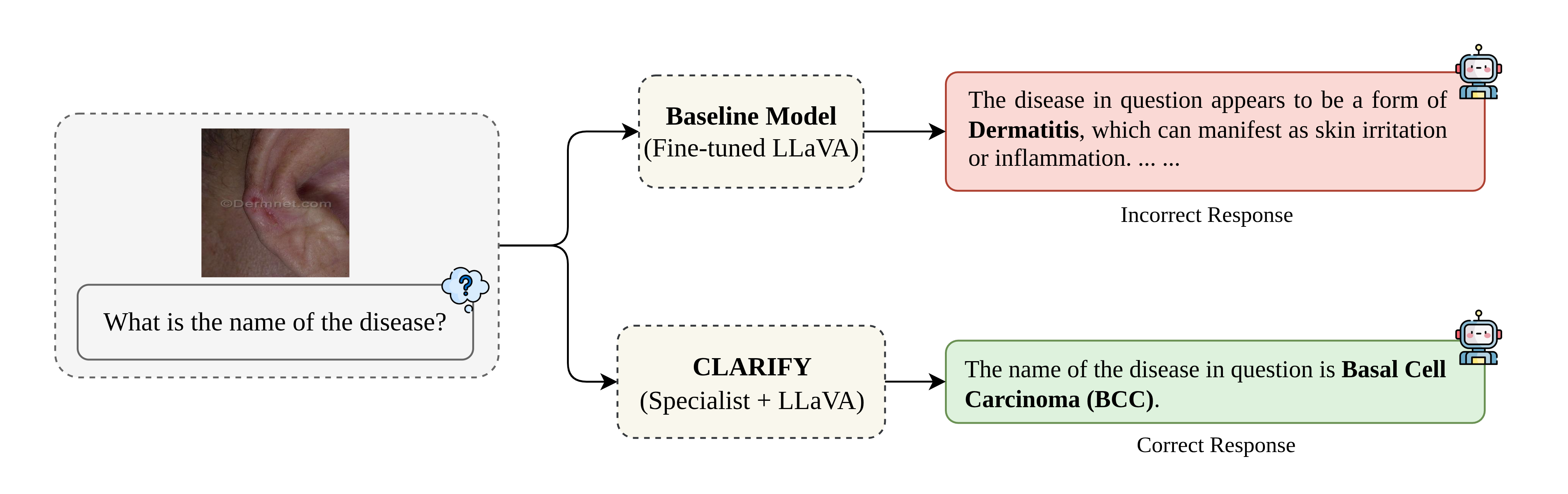}
  \caption{Qualitative comparison on a sample query. The Baseline VLM (left) hallucinates an incorrect diagnosis. The CLARIFY system (right), guided by the Specialist's correct prediction (``Basal Cell Carcinoma (BCC)"), provides an accurate and reliable response.}
  \label{fig:clarify_qualitative}
\end{figure*}

\subsection{The CLARIFY Inference Pipeline: Integrating the Components}
At inference time, the three independently-prepared components are integrated into a sequential workflow that ensures accuracy, safety, and reliability. The global workflow is depicted in Figure~\ref{fig:dino_llava_kg}.

\begin{enumerate}
    \item \textbf{Perception by the Specialist:} The user submits an image and a textual query. The image is first processed by the \textbf{Specialist} module, which outputs a predicted disease class.

    \item \textbf{Knowledge Retrieval:} The predicted disease label is used to query the \textbf{Knowledge Module}. The system performs a semantic search on the pre-computed KG embeddings to find the most relevant entity. It then traverses the graph to retrieve related information (e.g., symptoms, common treatments, descriptions).

    \item \textbf{Guided Prompting for the Generalist:} The retrieved knowledge, the Specialist's diagnosis, and the user's original query are combined into a comprehensive prompt for the \textbf{Generalist}. The prompt is structured using a template that explicitly instructs the VLM to act as a medical assistant and synthesize the provided information as illustrated in Figure~\ref{fig:prompt}.

    \item \textbf{Grounded Response Generation:} The Generalist VLM receives the prompt and the original image, generating a final response that integrates the visual information with the diagnostic and factual context provided.
\end{enumerate}

This multi-stage pipeline ensures that the final output is anchored in an accurate visual classification and grounded in a verifiable knowledge base, directly addressing the core challenges of deploying reliable and trustworthy medical AI.


\section{Results and Analysis}\label{sec:results}

In this section, we present a comprehensive evaluation of the CLARIFY framework. Our experiments are designed to answer three key questions: (1) How much inference cost can be saved without significantly compromising the performance of the VLM backbone? (2) Does the integrated CLARIFY system substantially improve diagnostic accuracy compared to a conventional fine-tuned VLM? (3) Does the Knowledge Graph-based RAG component enhance the factual accuracy and relevance of the conversational responses?

\subsection{Evaluation Metric}\label{sub:eval}

For the diagnostic task, we measured classification accuracy (correctly classified images). For the conversational task, we employed an LLM-as-judge evaluation, using Qwen-2.5-32B and gpt-oss-20b to assess response quality against a ground-truth knowledge base.

To demonstrate the efficacy of our framework, we compare CLARIFY against a strong baseline: a standard end-to-end fine-tuned VLM. This baseline model undergoes a two-stage fine-tuning process (first on image-label pairs for diagnosis, then on conversational data), representing the conventional approach to specializing a VLM without a decoupled architecture. In our experiment, we fine-tuned three models using this two-stage process to create our baseline models: Qwen2.5-VL-3B, Qwen2.5-VL-7B, and LLaVA-1.5-7B.

\subsection{Performance of Pruned Model}

Structural pruning involves eliminating specific parameters, such as entire layers, from a neural network model to enhance efficiency. This process reduces the total number of parameters, leading to faster inference times and lower video random-access memory (VRAM) usage. However, this reduction in model complexity comes at the cost of diminished conversational performance, as removing components can impair the model's ability to generate accurate and coherent outputs.
To guide the pruning process, an importance score is calculated for each layer after iteratively removing individual blocks from the model. This score reflects the layer’s contribution to the model’s output accuracy: a higher score indicates that the model retains better performance even when that layer is removed, suggesting that the layer is less critical to the model’s functionality. Conversely, layers with lower importance scores are deemed more essential, as their removal significantly degrades performance.

Based on these importance scores, layers are systematically pruned, starting with those having the lowest cumulative importance (i.e., the least impact on performance when removed). This iterative pruning process allows for a controlled reduction in model size while monitoring the tradeoff between efficiency and performance. For performance measurement we use LLM-Based Judge\cite{gu2024survey}. We evaluate our compressed VLM generated text against the ground truth and provide a performance score from our judge LLMs.

The results of this structured pruning approach are summarized in Table~\ref{tab:structured_pruning_llm}, which illustrates the relationship between model compression, performance, and resource usage. Below is a detailed breakdown of the table:

\begin{table*}[h]
\resizebox{\linewidth}{!}{
    \centering
    \begin{tabular}{|l|c|c|c|c|c|c|c|}
        \hline
        \makecell{\textbf{Model Name}} & 
        \makecell{\textbf{Layers} \\ \textbf{Removed}} & 
        \makecell{\textbf{Number of} \\ \textbf{Parameters (B)}} & 
        \makecell{\textbf{Compression} \\ \textbf{Ratio (\%)}} & 
        \makecell{\textbf{VRAM (GB)}} & 
        \makecell{\textbf{Score, Judge:} \\ \textbf{Qwen-2.5-32B (\%)}} & 
        \makecell{\textbf{Score, Judge:} \\ \textbf{gpt-oss-20b (\%)}} &
        \makecell{\textbf{Latency} \\ \textbf{(ms/token)} } \\
        \hline
        \multirow{7}{*}{Qwen-2.5-3B}
                        & 0  & 3.750 & 0  & 7.2  & 76.5 & 79.5 & 144 \\
                        & 2  & 3.596 & 4  & 6.8  & 69.0 & 72.0 & 142 \\
                        & 4  & 3.442 & 8  & 6.4  & 61.0 & 63.5 & 140 \\
                        & 7  & 3.211 & 14 & 5.8  & 51.0 & 54.0 & 137 \\
                        & 8  & 3.134 & 16 & 5.6  & 43.0 & 45.5 & 135 \\
                        & 9  & 3.057 & 18 & 5.4  & 29.0 & 32.0 & 133 \\
                        & 10 & 2.980 & 21 & 5.2  & 21.0 & 23.5 & 132 \\
        \hline
        \multirow{7}{*}{LLaVA-1.5-7B} 
                        & 0  & 7.063 & 0  & 13.4 & 78.5 & 81.0 & 163 \\
                        & 2  & 6.659 & 5  & 12.5 & 71.5 & 74.0 & 159 \\
                        & 4  & 6.254 & 10 & 11.7 & 64.0 & 67.0 & 156 \\
                        & 7  & 5.647 & 20 & 10.5 & 56.0 & 59.0 & 153 \\
                        & 8  & 5.444 & 23 & 10.1 & 48.0 & 50.5 & 151 \\
                        & 9  & 5.242 & 26 & 9.7  & 35.0 & 38.0 & 149 \\
                        & 10 & 5.040 & 29 & 9.4  & 26.0 & 29.0 & 147 \\
        \hline
        \multirow{7}{*}{Qwen-2.5-7B}  
                        & 0  & 8.290 & 0  & 14.0 & 81.0 & 84.0 & 176 \\
                        & 2  & 7.824 & 6  & 13.1 & 75.0 & 78.0 & 173 \\
                        & 4  & 7.358 & 11 & 12.3 & 68.0 & 71.0 & 170 \\
                        & 7  & 6.659 & 20 & 11.1 & 59.0 & 62.0 & 165 \\
                        & 8  & 6.426 & 22 & 10.7 & 51.0 & 54.0 & 164 \\
                        & 9  & 6.193 & 25 & 10.3 & 38.0 & 41.0 & 162 \\
                        & 10 & 5.960 & 28 & 10.0 & 29.0 & 32.0 & 161 \\
        \hline
    \end{tabular}
}
    \caption{Results of Structured Pruning for LLaVA and Qwen Models with Evaluation from Two Judges}
    \label{tab:structured_pruning_llm}
\end{table*}

\subsection{End-to-End Diagnostic Accuracy: CLARIFY vs. Baseline}
We next evaluated the primary claim of our work: that the CLARIFY framework improves diagnostic accuracy. We compared the performance of our full CLARIFY system (Specialist + Generalist) against the baseline fine-tuned VLM across several model families.

To develop the Specialist Module, we fine-tuned DINOv2 (base variant, 86.6M parameters) on an image–label dataset derived solely from our curated Small-Derma-VQA Dataset until convergence. Table~\ref{tab:clarify_vs_baseline} compares the diagnostic accuracy of fine-tuned VLM baselines against our proposed CLARIFY framework. While Qwen-VL and LLaVA achieve moderate performance after fine-tuning, CLARIFY substantially outperforms all baselines, reaching an accuracy of 82.1\%. The diagnostic accuracy in CLARIFY framework is same for all the VLMs because the diagnosis comes from the specialist module. This result highlights the effectiveness of the Specialist–Generalist architecture in medical image understanding tasks, demonstrating clear advantages over standard end-to-end fine-tuning.

\begin{table}[h]
\centering
\begin{tabular}{|l|c|c|}
\hline
\textbf{Mode} & \textbf{Backbone Model} & \textbf{Accuracy (\%)} \\
\hline
\multirow{3}{*}{Fine-tuned} 
 & Qwen-VL-3B     & 51.3 \\
 & LLaVA-1.5-7B   & 53.8 \\
 & Qwen-VL-7B     & 64.1 \\
\hline
\multirow{2}{*}{CLARIFY} 
 & Specialist (DINOv2) + & \multirow{2}{*}{\textbf{82.1}} \\
 & Generalist (Qwen/LLaVA) & \\
\hline
\end{tabular}
\caption{End-to-end diagnostic accuracy comparison across models. CLARIFY, which integrates both specialist and generalist modules, achieves the highest performance.}
\label{tab:clarify_vs_baseline}
\end{table}

The qualitative difference is stark, as shown in Figure~\ref{fig:clarify_qualitative}. The baseline model, confused by conversational fine-tuning, misidentifies the disease. In contrast, the CLARIFY system, guided by the Specialist's correct prediction, provides the right diagnosis along with a grounded explanation.

\subsection{Conversational Quality and Factual Grounding}
Finally, we evaluated the quality of the conversational responses generated by the full CLARIFY system, which leverages the KG-RAG module. We used the LLM-as-a-Judge framework to assess the similarity and accuracy of the models’ responses to questions about treatment, causes, and related topics against ground-truth information. For the final evaluation of CLARIFY, we used pruned VLMs with seven transformer layers removed as the Generalist (as shown in Table~\ref{tab:structured_pruning_llm}). This configuration was chosen because it provides a substantial reduction in VRAM usage without significantly compromising performance.

As mentioned earlier in Section~\ref{sub:eval}, the baseline models were created using two-stage fine-tuning. However, this two-stage process resulted in a noticeable drop in general conversational ability. In contrast, CLARIFY employed a single-stage fine-tuning approach focused solely on conversational adaptation of the language model, while keeping the vision encoder frozen. This approach explains why the final CLARIFY framework demonstrates strong conversational performance.

\begin{table}[h]
    \centering
    \resizebox{\linewidth}{!}{%
        \begin{tabular}{|l|l|c|c|}
            \hline
            \textbf{Mode} & \textbf{Backbone Model} & \makecell{\textbf{Score, Judge:} \\ \textbf{Qwen-2.5-32B (\%)}} & \makecell{\textbf{Score, Judge:} \\ \textbf{gpt-oss-20b (\%)}} \\
            \hline
            \multirow{3}{*}{Fine-Tuned} 
                & Qwen-VL-3B & 60 & 65 \\
                & LLaVA-1.5-7B & 66 & 69 \\
                & Qwen-VL-7B & 71 & 78 \\
            \hline
            \multirow{3}{*}{CLARIFY} 
                & Qwen-VL-3B & 75 & 83 \\
                & LLaVA-1.5-7B & 79 & 84 \\
                & Qwen-VL-7B & 84 & 89 \\
            \hline
        \end{tabular}
    }
    \caption{Overall LLM-as-a-Judge evaluation of conversational quality (single combined score per model, scaled to 0–100). CLARIFY achieves consistently higher performance across all backbones, with CLARIFY (Qwen-VL-7B) being the strongest. The Specialist Module is not explicitly shown here but is included within the CLARIFY framework.}
    \label{tab:rag_evaluation}
\end{table}

As shown in Table~\ref{tab:rag_evaluation}, the CLARIFY system with various Generalist backbones consistently produced high-quality responses. Both Qwen-2.5-32B and gpt-oss-20b assigned high scores for similarity and accuracy across different topics, indicating that the KG-RAG module successfully grounds the Generalist's outputs in factual medical knowledge. The consistent scores suggest a robust and reliable level of performance, ensuring that the conversational aspect of the system is both helpful and trustworthy.

In summary, our results demonstrate that the CLARIFY framework is highly effective. It significantly boosts diagnostic accuracy by decoupling perception and reasoning, while ensuring that its conversational outputs are factually grounded and reliable through its integrated KG-RAG module.

\section{Conclusion}\label{sec:conclusion}

In this work, we introduced \textbf{CLARIFY}, a novel \textbf{Specialist–Generalist framework} designed to resolve the tension between diagnostic accuracy and conversational fluency in medical AI. We argued that forcing a single, monolithic Vision–Language Model (VLM) to handle both high-stakes classification and nuanced dialogue leads to catastrophic forgetting and unreliable outputs. To address this, CLARIFY decouples perception and reasoning: a lightweight, expert \textbf{Specialist} classifier performs disease recognition, while a compressed, conversationally tuned \textbf{Generalist} VLM generates coherent and safe responses, guided by the Specialist’s output and grounded in a Knowledge Graph–based retrieval module. Our experiments demonstrate that CLARIFY substantially outperforms conventional fine-tuned VLMs, achieving an accuracy of 82.1\% compared to the 51–64\% range of baselines. Beyond accuracy, the integrated KG-RAG component ensures that responses remain factually grounded, improving both reliability and trustworthiness in clinical settings. These results validate the effectiveness of separating perception from reasoning when designing medical AI systems. Despite its promise, CLARIFY has limitations that motivate future work. Its multi-component design introduces architectural complexity, and the overall performance depends on the breadth and quality of the underlying Knowledge Graph. Future research will focus on (1) applying advanced compression to further minimize the Generalist’s footprint, (2) expanding and validating the Knowledge Graph with expert input, and (3) conducting prospective clinical studies to assess real-world utility and safety. In summary, CLARIFY provides a robust and practical paradigm for medical AI: combining the precision of expert classifiers with the reasoning capabilities of large language models. This Specialist–Generalist blueprint offers a path toward AI systems that are accurate, efficient, interpretable, and ultimately ready for deployment in dermatology and other high-stakes medical domains.

\section{Limitations and Future Work}
Our experiments were conducted on a relatively small dataset due to resource constraints, which limits the generalizability and robustness of the results. Scaling to larger and more diverse datasets is essential to fully evaluate the framework's performance across a broader range of dermatological conditions. Additionally, the knowledge graph used in this work was derived from Wikipedia for experimental demonstration, and we acknowledge it is not a clinically authoritative source. Future work will involve training and testing with substantially larger datasets and incorporating clinically validated ontologies, along with rigorous validation by medical experts to ensure clinical reliability and applicability.

\section{Acknowledgements}
During the preparation of this work, the authors used AI-based writing assistance tools to improve language clarity, check grammar, and refine sentence structure. The authors take full responsibility for the final content and all intellectual contributions of this paper.

\bibliographystyle{IEEEtran}
\bibliography{IEEEabrv,bibliography}

\begin{thebibliography}{10}
\providecommand{\url}[1]{#1}
\csname url@samestyle\endcsname
\providecommand{\newblock}{\relax}
\providecommand{\bibinfo}[2]{#2}
\providecommand{\BIBentrySTDinterwordspacing}{\spaceskip=0pt\relax}
\providecommand{\BIBentryALTinterwordstretchfactor}{4}
\providecommand{\BIBentryALTinterwordspacing}{\spaceskip=\fontdimen2\font plus
\BIBentryALTinterwordstretchfactor\fontdimen3\font minus \fontdimen4\font\relax}
\providecommand{\BIBforeignlanguage}[2]{{%
\expandafter\ifx\csname l@#1\endcsname\relax
\typeout{** WARNING: IEEEtran.bst: No hyphenation pattern has been}%
\typeout{** loaded for the language `#1'. Using the pattern for}%
\typeout{** the default language instead.}%
\else
\language=\csname l@#1\endcsname
\fi
#2}}
\providecommand{\BIBdecl}{\relax}
\BIBdecl

\bibitem{liu2023visualinstructiontuning}
\BIBentryALTinterwordspacing
H.~Liu, C.~Li, Q.~Wu, and Y.~J. Lee, ``Visual instruction tuning,'' 2023. [Online]. Available: \url{https://arxiv.org/abs/2304.08485}
\BIBentrySTDinterwordspacing

\bibitem{bai2023qwenvlversatilevisionlanguagemodel}
\BIBentryALTinterwordspacing
J.~Bai, S.~Bai, S.~Yang, S.~Wang, S.~Tan, P.~Wang, J.~Lin, C.~Zhou, and J.~Zhou, ``Qwen-vl: A versatile vision-language model for understanding, localization, text reading, and beyond,'' 2023. [Online]. Available: \url{https://arxiv.org/abs/2308.12966}
\BIBentrySTDinterwordspacing

\bibitem{zhang2024vision}
J.~Zhang, J.~Huang, S.~Jin, and S.~Lu, ``Vision-language models for vision tasks: A survey,'' \emph{IEEE transactions on pattern analysis and machine intelligence}, vol.~46, no.~8, pp. 5625--5644, 2024.

\bibitem{chen2023medblipbootstrappinglanguageimagepretraining}
\BIBentryALTinterwordspacing
Q.~Chen, X.~Hu, Z.~Wang, and Y.~Hong, ``Medblip: Bootstrapping language-image pre-training from 3d medical images and texts,'' 2023. [Online]. Available: \url{https://arxiv.org/abs/2305.10799}
\BIBentrySTDinterwordspacing

\bibitem{tu2024towards}
T.~Tu, S.~Azizi, D.~Driess, M.~Schaekermann, M.~Amin, P.-C. Chang, A.~Carroll, C.~Lau, R.~Tanno, I.~Ktena \emph{et~al.}, ``Towards generalist biomedical ai,'' \emph{Nejm Ai}, vol.~1, no.~3, p. AIoa2300138, 2024.

\bibitem{Wu2023MultimodalLL}
\BIBentryALTinterwordspacing
J.~Wu, W.~Gan, Z.~Chen, S.~Wan, and P.~S. Yu, ``Multimodal large language models: A survey,'' \emph{2023 IEEE International Conference on Big Data (BigData)}, pp. 2247--2256, 2023. [Online]. Available: \url{https://api.semanticscholar.org/CorpusID:265351653}
\BIBentrySTDinterwordspacing

\bibitem{Santomauro2023EnhancingMI}
\BIBentryALTinterwordspacing
A.~Santomauro, L.~Portinale, and G.~Leonardi, ``Enhancing medical image report generation through standard language models: Leveraging the power of llms in healthcare,'' in \emph{HC@AIxIA}, 2023. [Online]. Available: \url{https://api.semanticscholar.org/CorpusID:266211540}
\BIBentrySTDinterwordspacing

\bibitem{doi:10.1073/pnas.1611835114}
\BIBentryALTinterwordspacing
J.~Kirkpatrick, R.~Pascanu, N.~Rabinowitz, J.~Veness, G.~Desjardins, A.~A. Rusu, K.~Milan, J.~Quan, T.~Ramalho, A.~Grabska-Barwinska, D.~Hassabis, C.~Clopath, D.~Kumaran, and R.~Hadsell, ``Overcoming catastrophic forgetting in neural networks,'' \emph{Proceedings of the National Academy of Sciences}, vol. 114, no.~13, pp. 3521--3526, 2017. [Online]. Available: \url{https://www.pnas.org/doi/abs/10.1073/pnas.1611835114}
\BIBentrySTDinterwordspacing

\bibitem{luo2025empiricalstudycatastrophicforgetting}
\BIBentryALTinterwordspacing
Y.~Luo, Z.~Yang, F.~Meng, Y.~Li, J.~Zhou, and Y.~Zhang, ``An empirical study of catastrophic forgetting in large language models during continual fine-tuning,'' 2025. [Online]. Available: \url{https://arxiv.org/abs/2308.08747}
\BIBentrySTDinterwordspacing

\bibitem{sun2024simpleeffectivepruningapproach}
\BIBentryALTinterwordspacing
M.~Sun, Z.~Liu, A.~Bair, and J.~Z. Kolter, ``A simple and effective pruning approach for large language models,'' 2024. [Online]. Available: \url{https://arxiv.org/abs/2306.11695}
\BIBentrySTDinterwordspacing

\bibitem{ma2023llmprunerstructuralpruninglarge}
\BIBentryALTinterwordspacing
X.~Ma, G.~Fang, and X.~Wang, ``Llm-pruner: On the structural pruning of large language models,'' 2023. [Online]. Available: \url{https://arxiv.org/abs/2305.11627}
\BIBentrySTDinterwordspacing

\bibitem{frantar2023gptqaccurateposttrainingquantization}
\BIBentryALTinterwordspacing
E.~Frantar, S.~Ashkboos, T.~Hoefler, and D.~Alistarh, ``Gptq: Accurate post-training quantization for generative pre-trained transformers,'' 2023. [Online]. Available: \url{https://arxiv.org/abs/2210.17323}
\BIBentrySTDinterwordspacing

\bibitem{lang2024comprehensivestudyquantizationtechniques}
\BIBentryALTinterwordspacing
J.~Lang, Z.~Guo, and S.~Huang, ``A comprehensive study on quantization techniques for large language models,'' 2024. [Online]. Available: \url{https://arxiv.org/abs/2411.02530}
\BIBentrySTDinterwordspacing

\bibitem{lewis2021retrievalaugmentedgenerationknowledgeintensivenlp}
\BIBentryALTinterwordspacing
P.~Lewis, E.~Perez, A.~Piktus, F.~Petroni, V.~Karpukhin, N.~Goyal, H.~Küttler, M.~Lewis, W.~tau Yih, T.~Rocktäschel, S.~Riedel, and D.~Kiela, ``Retrieval-augmented generation for knowledge-intensive nlp tasks,'' 2021. [Online]. Available: \url{https://arxiv.org/abs/2005.11401}
\BIBentrySTDinterwordspacing

\bibitem{Pan_2024}
\BIBentryALTinterwordspacing
S.~Pan, L.~Luo, Y.~Wang, C.~Chen, J.~Wang, and X.~Wu, ``Unifying large language models and knowledge graphs: A roadmap,'' \emph{IEEE Transactions on Knowledge and Data Engineering}, vol.~36, no.~7, p. 3580–3599, Jul. 2024. [Online]. Available: \url{http://dx.doi.org/10.1109/TKDE.2024.3352100}
\BIBentrySTDinterwordspacing

\bibitem{xiong2024improvingretrievalaugmentedgenerationmedicine}
\BIBentryALTinterwordspacing
G.~Xiong, Q.~Jin, X.~Wang, M.~Zhang, Z.~Lu, and A.~Zhang, ``Improving retrieval-augmented generation in medicine with iterative follow-up questions,'' 2024. [Online]. Available: \url{https://arxiv.org/abs/2408.00727}
\BIBentrySTDinterwordspacing

\bibitem{dermnet2025}
{DermNet}, ``Dermatology resource,'' \url{https://dermnetnz.org}, 2025, accessed: 2025-08-19.

\bibitem{radford2021learningtransferablevisualmodels}
\BIBentryALTinterwordspacing
A.~Radford, J.~W. Kim, C.~Hallacy, A.~Ramesh, G.~Goh, S.~Agarwal, G.~Sastry, A.~Askell, P.~Mishkin, J.~Clark, G.~Krueger, and I.~Sutskever, ``Learning transferable visual models from natural language supervision,'' 2021. [Online]. Available: \url{https://arxiv.org/abs/2103.00020}
\BIBentrySTDinterwordspacing

\bibitem{li2022blipbootstrappinglanguageimagepretraining}
\BIBentryALTinterwordspacing
J.~Li, D.~Li, C.~Xiong, and S.~Hoi, ``Blip: Bootstrapping language-image pre-training for unified vision-language understanding and generation,'' 2022. [Online]. Available: \url{https://arxiv.org/abs/2201.12086}
\BIBentrySTDinterwordspacing

\bibitem{he2015deepresiduallearningimage}
\BIBentryALTinterwordspacing
K.~He, X.~Zhang, S.~Ren, and J.~Sun, ``Deep residual learning for image recognition,'' 2015. [Online]. Available: \url{https://arxiv.org/abs/1512.03385}
\BIBentrySTDinterwordspacing

\bibitem{dosovitskiy2021imageworth16x16words}
\BIBentryALTinterwordspacing
A.~Dosovitskiy, L.~Beyer, A.~Kolesnikov, D.~Weissenborn, X.~Zhai, T.~Unterthiner, M.~Dehghani, M.~Minderer, G.~Heigold, S.~Gelly, J.~Uszkoreit, and N.~Houlsby, ``An image is worth 16x16 words: Transformers for image recognition at scale,'' 2021. [Online]. Available: \url{https://arxiv.org/abs/2010.11929}
\BIBentrySTDinterwordspacing

\bibitem{esteva2017dermatologist}
A.~Esteva, B.~Kuprel, R.~A. Novoa, J.~Ko, S.~M. Swetter, H.~M. Blau, and S.~Thrun, ``Dermatologist-level classification of skin cancer with deep neural networks,'' \emph{nature}, vol. 542, no. 7639, pp. 115--118, 2017.

\bibitem{xu2023survey}
C.~Xu and J.~McAuley, ``A survey on model compression and acceleration for pretrained language models,'' in \emph{Proceedings of the AAAI Conference on Artificial Intelligence}, vol.~37, no.~9, 2023, pp. 10\,566--10\,575.

\bibitem{kim2024shortened}
B.-K. Kim, G.~Kim, T.-H. Kim, T.~Castells, S.~Choi, J.~Shin, and H.-K. Song, ``Shortened llama: A simple depth pruning for large language models,'' \emph{arXiv preprint arXiv:2402.02834}, vol.~11, 2024.

\bibitem{men2024shortgptlayerslargelanguage}
\BIBentryALTinterwordspacing
X.~Men, M.~Xu, Q.~Zhang, B.~Wang, H.~Lin, Y.~Lu, X.~Han, and W.~Chen, ``Shortgpt: Layers in large language models are more redundant than you expect,'' 2024. [Online]. Available: \url{https://arxiv.org/abs/2403.03853}
\BIBentrySTDinterwordspacing

\bibitem{zhao2025medragenhancingretrievalaugmentedgeneration}
\BIBentryALTinterwordspacing
X.~Zhao, S.~Liu, S.-Y. Yang, and C.~Miao, ``Medrag: Enhancing retrieval-augmented generation with knowledge graph-elicited reasoning for healthcare copilot,'' 2025. [Online]. Available: \url{https://arxiv.org/abs/2502.04413}
\BIBentrySTDinterwordspacing

\bibitem{sanmartin2024kgragbridginggapknowledge}
\BIBentryALTinterwordspacing
D.~Sanmartin, ``Kg-rag: Bridging the gap between knowledge and creativity,'' 2024. [Online]. Available: \url{https://arxiv.org/abs/2405.12035}
\BIBentrySTDinterwordspacing

\bibitem{gregg2022dedupe}
F.~Gregg and D.~Eder, ``Dedupe,'' \url{https://github.com/dedupeio/dedupe}, 2022, accessed: 2025-08-19.

\bibitem{aranya_saha_tanvir_ahmed_khan_ismam_nur_swapnil_2025}
\BIBentryALTinterwordspacing
A.~Saha, T.~A. Khan, and I.~N. Swapnil, ``Small-derma-vqa,'' 2025. [Online]. Available: \url{https://www.kaggle.com/dsv/12845315}
\BIBentrySTDinterwordspacing

\bibitem{oquab2023dinov2}
M.~Oquab, T.~Darcet, T.~Moutakanni, H.~Vo, M.~Szafraniec, V.~Khalidov, P.~Fernandez, D.~Haziza, F.~Massa, A.~El-Nouby \emph{et~al.}, ``Dinov2: Learning robust visual features without supervision,'' \emph{arXiv preprint arXiv:2304.07193}, 2023.

\bibitem{ma2023llm}
X.~Ma, G.~Fang, and X.~Wang, ``Llm-pruner: On the structural pruning of large language models,'' \emph{Advances in neural information processing systems}, vol.~36, pp. 21\,702--21\,720, 2023.

\bibitem{mo2025kggen}
B.~Mo, K.~Yu, J.~Kazdan, P.~Mpala, L.~Yu, C.~Cundy, C.~Kanatsoulis, and S.~Koyejo, ``Kggen: Extracting knowledge graphs from plain text with language models,'' \emph{arXiv preprint arXiv:2502.09956}, 2025.

\bibitem{geminiteam2025geminifamilyhighlycapable}
G.~Team, R.~Anil, S.~Borgeaud, J.-B. Alayrac, J.~Yu, R.~Soricut, J.~Schalkwyk, A.~M. Dai, A.~Hauth, K.~Millican \emph{et~al.}, ``Gemini: a family of highly capable multimodal models,'' \emph{arXiv preprint arXiv:2312.11805}, 2023.

\bibitem{khattab2023dspy}
O.~Khattab, A.~Singhvi, P.~Maheshwari, Z.~Zhang, K.~Santhanam, S.~Vardhamanan, S.~Haq, A.~Sharma, T.~T. Joshi, H.~Moazam \emph{et~al.}, ``Dspy: Compiling declarative language model calls into self-improving pipelines,'' \emph{arXiv preprint arXiv:2310.03714}, 2023.

\bibitem{wang2020minilmdeepselfattentiondistillation}
\BIBentryALTinterwordspacing
W.~Wang, F.~Wei, L.~Dong, H.~Bao, N.~Yang, and M.~Zhou, ``Minilm: Deep self-attention distillation for task-agnostic compression of pre-trained transformers,'' 2020. [Online]. Available: \url{https://arxiv.org/abs/2002.10957}
\BIBentrySTDinterwordspacing

\bibitem{gu2024survey}
J.~Gu, X.~Jiang, Z.~Shi, H.~Tan, X.~Zhai, C.~Xu, W.~Li, Y.~Shen, S.~Ma, H.~Liu \emph{et~al.}, ``A survey on llm-as-a-judge,'' \emph{arXiv preprint arXiv:2411.15594}, 2024.

\end{thebibliography}

\newpage
\section{Biography Section}
\vspace{-20pt}
\begin{IEEEbiography}[{\includegraphics[width=1in,height=1in, keepaspectratio]{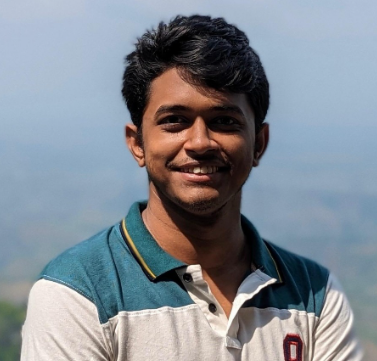}}]{Aranya Saha} received the B.Sc. degree in Electrical and Electronic Engineering, majoring in Communication and Signal Processing, from Bangladesh University of Engineering and Technology (BUET), Dhaka, Bangladesh, in 2025, where he is currently pursuing the M.Sc. degree in the same department and major. He is also working as a Machine Learning Engineer at Advanced Chemical Industries (ACI) Ltd., Dhaka. From 2024 to 2025, he served as a Student Executive of ACM SIGCOMM, contributing to community-building and technical initiatives for the global networking research community. His research interests include vision-language models, multimodal learning, computer vision, and autonomous systems.
\end{IEEEbiography}

\begin{IEEEbiography}[{\includegraphics[width=1in,height=1in]{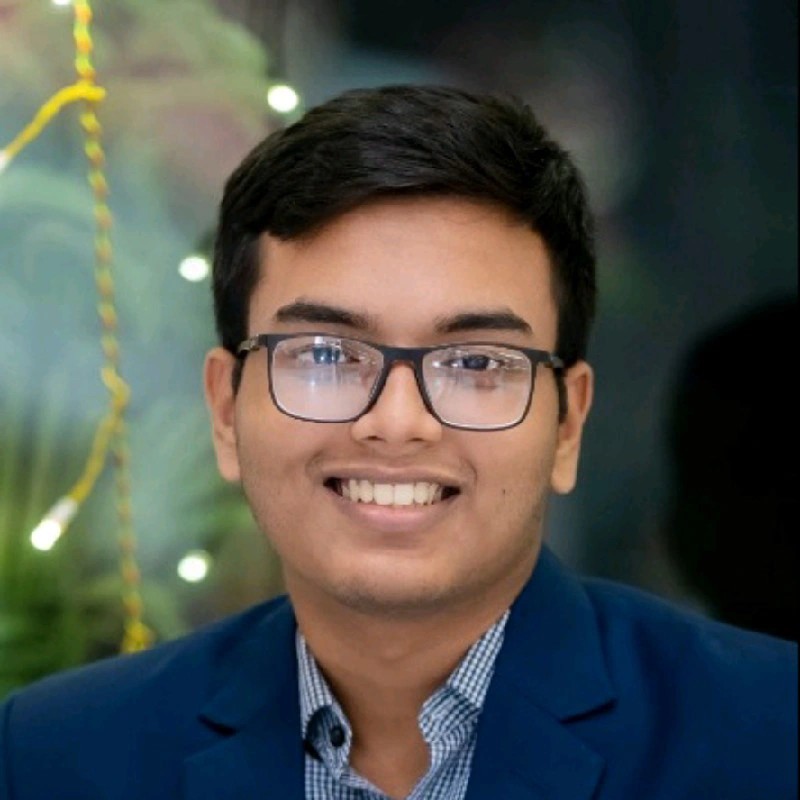}}]{Tanvir Ahmed Khan} received the B.Sc. degree in Electrical and Electronic Engineering, majoring in Communication and Signal Processing, from Bangladesh University of Engineering and Technology (BUET), Dhaka, Bangladesh, in 2025. His research interests include vision language models, multimodal learning, model compression, and model optimization.
\end{IEEEbiography}

\begin{IEEEbiography}[{\includegraphics[width=1in,height=1in]{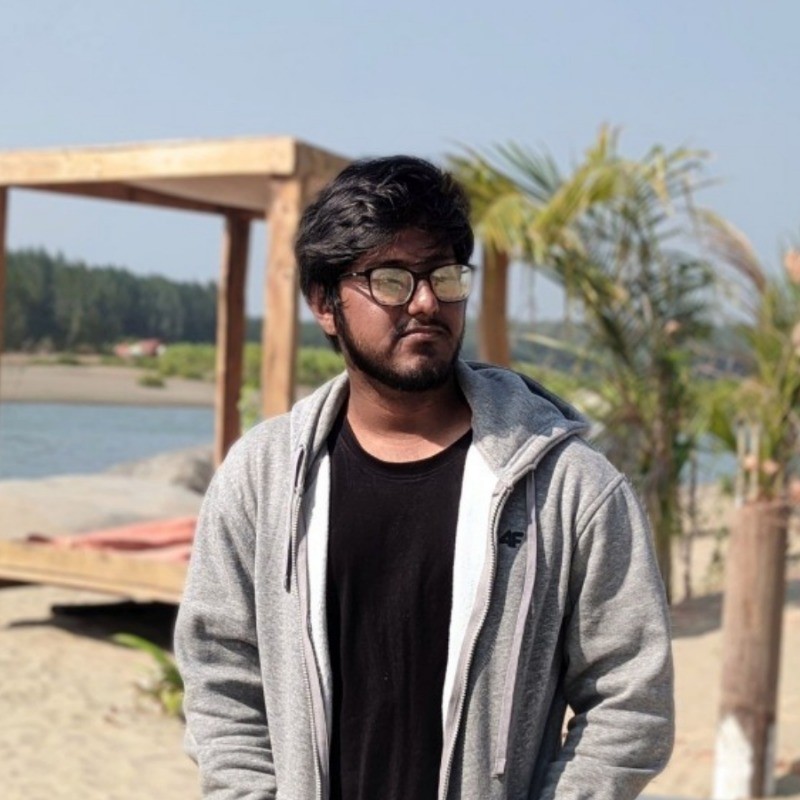}}]{Ismam Nur Swapnil} received the B.Sc. degree in Electrical and Electronic Engineering, majoring in Communication and Signal Processing, from Bangladesh University of Engineering and Technology (BUET), Dhaka, Bangladesh, in 2025. He is currently working as a Machine Learning Engineer at Advanced Chemical Industries (ACI) Ltd., Dhaka. His research interests include large language models, multimodal learning, model reasoning, and model optimization.
\end{IEEEbiography}

\begin{IEEEbiography}[{\includegraphics[width=1in,height=1in,clip]{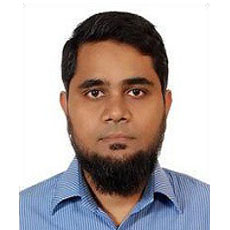}}]{Mohammad Ariful Haque}
(Member, IEEE) received the B.Sc., M.Sc., and Ph.D. degrees in Electrical and Electronic Engineering from Bangladesh University of Engineering and Technology (BUET), Dhaka, Bangladesh, in 2003, 2005, and 2009, respectively. He was a Post-Doctoral Fellow at Concordia University, Montreal, QC, Canada. In 2003, he joined the Department of Electrical and Electronic Engineering, BUET, as a Lecturer, where he is currently a Professor. Prof. Haque received the ReSMiQ Post-Doctoral Fellowship Award in 2012. His research interests include digital signal processing, deep learning, speech recognition, and biomedical signal processing. He has supervised finalist teams in global competitions organized by the IEEE Signal Processing Society at the IEEE International Conference on Acoustics, Speech, and Signal Processing (ICASSP) and the IEEE International Conference on Image Processing (ICIP) from 2014 to 2022, for nine consecutive years.
\end{IEEEbiography}

\vfill

\end{document}